\documentclass[journal]{IEEEtran}
\usepackage[colorlinks=true]{hyperref}
\usepackage{hyperref}
\usepackage{graphicx}
\usepackage{amsmath,amssymb}
\usepackage{color}
\usepackage{xcolor}
\usepackage[font=small,labelfont=bf]{caption}
\usepackage{makeidx}  
\usepackage{color}
\usepackage{graphicx}
\usepackage{amsmath}
\usepackage{amssymb}
\usepackage[labelformat=simple]{subfig}
\usepackage[ruled,lined,linesnumbered]{algorithm2e}
\usepackage{algorithmic}
\usepackage{blindtext}
\usepackage{enumitem}
\usepackage{multirow}
\usepackage[labelformat=simple]{subfig}
\usepackage{threeparttable/threeparttable}
\usepackage{bm}
\usepackage[normalem]{ulem}
\usepackage{hyperref}
\usepackage[utf8]{inputenc}
\usepackage{bibunits}
\usepackage{colortbl}
\usepackage{amssymb}
\usepackage{pifont}
\usepackage{array}
\usepackage{booktabs}
\usepackage{setspace}

\definecolor{maroon}{cmyk}{0,0.87,0.68,0.32}
\definecolor{brown(traditional)}{rgb}{0.59, 0.29, 0.0}

\newcolumntype{P}[1]{>{\centering\arraybackslash}p{#1}}

\newenvironment{jlrev}{}{}

\newenvironment{jlblue}{\color{black}}{}

\definecolor{zlue}{rgb}{0.22, 0.09, 1}

\usepackage{orcidlink}
\usepackage{authblk}

\begin{document}

\title{RotCAtt-TransUNet++: Novel Deep Neural Network for Sophisticated Cardiac Segmentation}

\author[1,2,3]{Quoc-Bao Nguyen-Le \orcidlink{0009-0001-4685-949X}}
\author[1]{Tuan-Hy Le \orcidlink{0009-0002-6290-3494}}
\author[1]{Anh-Triet Do \orcidlink{0009-0008-6475-6786}}
\author[2, 3]{Quoc-Huy Trinh \orcidlink{0000-0002-7205-3211}}

\affil[1]{Le Hong Phong High School for the Gifted, Ho Chi Minh City, Vietnam}
\affil[2]{Faculty of Information Technology, University of Science, VNU-HCM, Ho Chi Minh City, Vietnam}
\affil[3]{Viet Nam National University, Ho Chi Minh City, Vietnam}

\maketitle

\begin{abstract}
Cardiovascular disease remains a predominant global health concern, responsible for a significant portion of mortality worldwide. Accurate segmentation of cardiac medical imaging data is pivotal in mitigating fatality rates associated with cardiovascular conditions. However, existing state-of-the-art (SOTA) neural networks, including both CNN-based and Transformer-based approaches, exhibit limitations in practical applicability due to their inability to effectively capture inter-slice connections alongside intra-slice information. This deficiency is particularly evident in datasets featuring intricate, long-range details along the z-axis, such as coronary arteries in axial views. Additionally, SOTA methods fail to differentiate non-cardiac components from myocardium in segmentation, leading to the "spraying" phenomenon. To address these challenges, we present RotCAtt-TransUNet++, a novel architecture tailored for robust segmentation of complex cardiac structures. Our approach emphasizes modeling global contexts by aggregating multiscale features with nested skip connections in the encoder. It integrates transformer layers to capture interactions between patches (intra-slice information) and employs a rotatory attention mechanism to capture connectivity between multiple slices (inter-slice information). Additionally, a channel-wise cross-attention gate guides the fused multi-scale channel-wise information and features from decoder stages to bridge semantic gaps. Experimental results demonstrate that our proposed model outperforms existing SOTA approaches across four cardiac datasets and one abdominal dataset. Importantly, coronary arteries and myocardium are annotated with near-perfect accuracy during inference. An ablation study shows that the rotatory attention mechanism effectively transforms embedded vectorized patches in the semantic dimensional space, enhancing segmentation accuracy, thus offering better assistance for medical health industry.
\end{abstract}

\section{Introduction}
\label{sec:introduction}
Medical image segmentation plays a pivotal role in the detection of various diseases and tumors, offering accurate delineation of anatomical structures for enhanced visualization and analysis, particularly in 3D reconstructions of multiple internal organs. Significant advancements have been made across various medical domains, including cardiac segmentation from magnetic resonance (MR) imaging \cite{phi_vu_tran}, computed tomography (CT) scans \cite{contour}, and polyp segmentation from colonoscopy videos \cite{unet_plusplus}. While manual segmentation remains the gold standard in delineating pathological structures, it is labor-intensive, time-consuming, and reliant on expert knowledge, making it susceptible to human error \cite{transnorm}. Consequently, there is a growing interest in automated medical image segmentation, aimed at alleviating the burden of manual annotation.

Previous studies have primarily relied on single-labeled datasets such as the Sunnybrook Cardiac Data (SCD) from the 2009 Cardiac MR Left Ventricle Segmentation Challenge \cite{left_ventricle}, STACOM (2011) \cite{2011_left}, and the MICCAI Right Ventricle dataset (2012) \cite{right_ventricle}. However, recent advancements have introduced numerous 2D networks evaluated on multi-labeled cardiac datasets like the Multi-Modality Whole Heart Segmentation (MMWHS) from 2017 and the Automated Cardiac Diagnosis Challenge Dataset (ACDC) also from 2017. Nevertheless, these datasets typically only annotate basic regions: ACDC labels the right ventricle (RV), left ventricle (LV), and myocardium (Myo), while MMWHS includes seven fundamental regions but lacks significant details such as coronary arteries and cardiac capillaries. However, there are two other more complex datasets (e.g., ImageCHD, VHSCDD) that are less popular but challenge state-of-the-art (SOTA) methods. Detailed analysis by radiologists will benefit significantly from these sophisticated datasets, making highly accurate segmentation on them essential.

\begin{figure}[t]
\begin{center}
\includegraphics[width=1.0\linewidth]{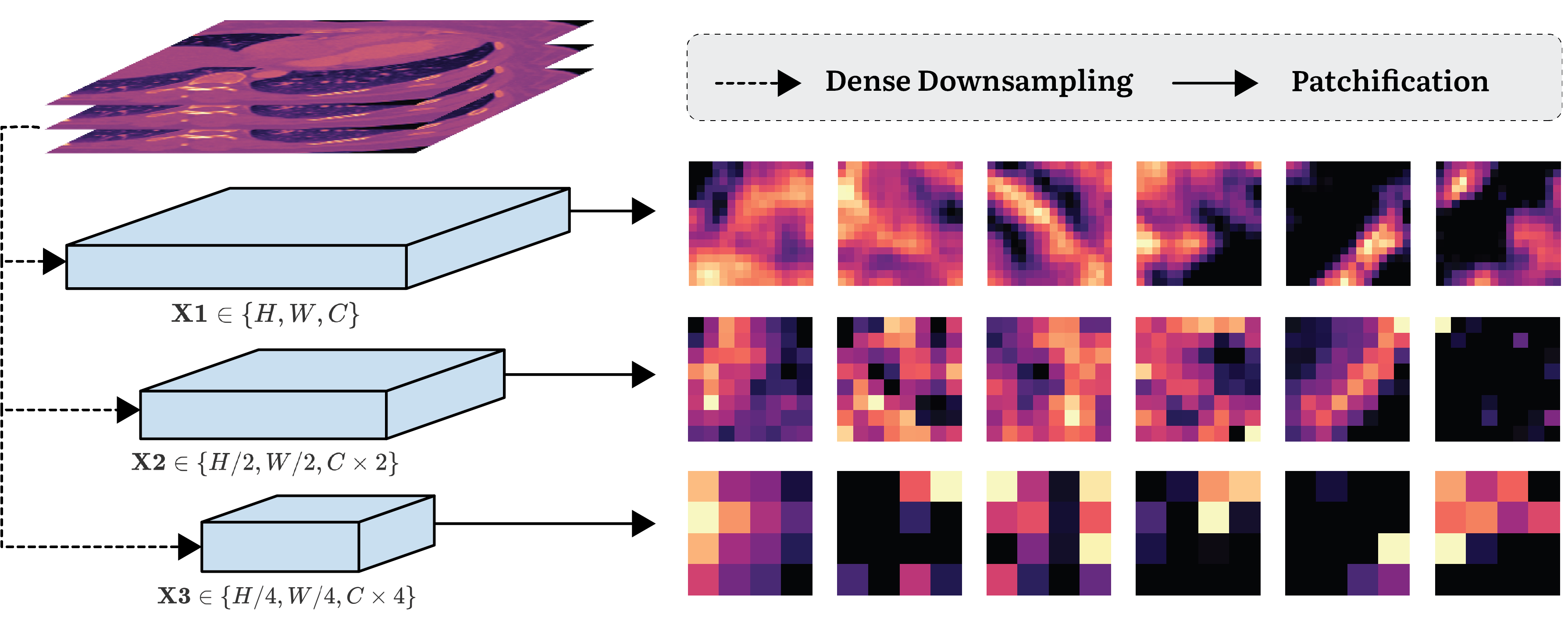}
\end{center}
\caption{Visualization of multi-scale feature maps after dense downsampling. The multi-scale learning enables the model to capture high-level features while preserving spatial information. Patches are depicted solely on the first feature map of $X1, X2, X3$ following convolutional operations and dense skip connections.}
\label{fig:predict_visualization}
\end{figure}

The current state-of-the-art 2D networks, including TransUNet, Swin-Unet, V-Net, ResUNet, UNet++, UNetR, and 3D Bidirectional Transformer Unet, have not undergone evaluation using the same cardiac datasets. Notably, while Swin-Unet was assessed on the ACDC dataset, others were only tested on non-cardiac datasets such as Synapse, abdomen CT dataset, thorax CT dataset, BTCV, and MSD. This discrepancy leads to an unfair comparison of these networks in the realm of cardiac segmentation. Furthermore, there is a notable absence of research integrating the segmentation of coronaries arteries with other cardiac regions. Since models tend to overlook such intricate details, recent works often opt for performing coronary segmentation on CT scans as a binary task (distinguish between background and coronary arteries) to minimize distraction from other classes. This issue can be addressed straightforward by training two separate models: one specifically for coronary segmentation and one for remaining classes. However, the latter model still needs to produce pixel values for coronary regions, which are classified as different class. This complicates the process of combining the results from both models and conducting quantitative post-analysis tasks such as volume measurement.

In this paper, we proved that CNN-based methods inevitably have limitations in capturing long-range dependencies due to their inherited property of confined receptive field, thus inferior to Transformer-based approaches \cite{transunet}. We further proved that current SOTA networks either lack or does not have robust mechanism to capture and attend to interslice information. Our objective is to propose a novel network capable of effectively addressing all intricately labeled regions within cardiac structures, without disregarding essential details. Our ultimate aim is to achieve highly accurate segmentation across various cardiac datasets. The content of this paper is organized as follows. In Section~\ref{sec:RelatedWorks}, we briefly review existing methods related to our work. Then we present our proposed solution in Section~\ref{sec:Methođology}. Experiments and result analysis are in Section~\ref{sec:Experiments}. Finally, the conclusion and implication are in Section~\ref{sec:ResultsAndConclusion}.

\section{Related works}
\label{sec:RelatedWorks}

\subsection{Traditional methods}
Mathematical methodologies encompass cluster-based algorithms like K-means and active contour models reliant on local and global intensities \cite{general_attention}. Nonetheless, challenges such as variations in tissue appearance, low resolution, and indistinct boundaries undermine the robustness of these approaches against noise and diverse contrasts in medical imaging. Machine learning techniques, including model-based (e.g. active shape and appearance models) and atlas-based methods, have not exhibited superior efficacy in this domain as they frequently necessitate substantial feature engineering or pre-existing knowledge to attain acceptable accuracy \cite{deep_review}. More recently, Deep Learning (DL) techniques have emerged triumphant in various computer vision applications, including object recognition and semantic segmentation.

\subsection{Deep Learning methods}

\subsubsection{CNN-based approaches}
Convolutional neural networks (CNNs), particularly Fully Convolutional Neural Networks (FCNs), have become the de facto standard in medical image segmentation \cite{transnorm, transunet, fcn}, utilizing the U-shaped or encoder-decoder architecture. The encoder, responsible for downsampling to reduce spatial dimensions and capture hierarchical high-level features, while the decoder, responsible for upsampling, restores spatial details from the feature map back to the size of the input image. In 2016, Phi Vu Tran \cite{phi_vu_tran} applied this network for cardiac segmentation in short-axis MRI. However, these architectures face a significant challenge due to the loss of details in deeper layers of the network. To address this issue, UNet were devised, notable with notable with direct skip connections that join feature maps at the same scale. This is one of the earliest and most widely used techniques in medical image segmentation, was developed by Ronneberger et al. based on the encoder-decoder architecture. Originally employed for Electron Microscopy Image (EM) segmentation in the International Symposium on Biomedical Imaging (ISBI) 2012 challenge. However, U-Net has several shortcomings, including direct skip connections that join feature maps from the same scale without considering the relationship between feature maps from different stages, leading to a large semantic gap problem. U-Net++ \cite{unet_plusplus} addresses this by employing nested or dense skip connections between different stages using various shortcut connections to reduce the semantic gap between encoder and decoder, aiming to capture deeper contextual representations. ResUNet \cite{resunet}, still based on encoder-decoder paradigm, replaces the standard convolutions with ResNet units that contain multiple in parallel atrous convolutions and pyramid pooling. Such modules boost algorithmic performance on semantic segmentation tasks and avoid vanishing gradients. However, it still suffers from a confined receptive field due to the nature of the convolution operation. Inherent inductive biases limits CNN-based technique from modeling long-range dependencies; pooling and convolution layers might prevent low-level features from being propagated to next convolutional layers. Above architectures generally yield weak performance especially for target structures that show large inter-patient variation in terms of texture, shape, and size \cite{transunet}.

Various studies have attempted to integrate self-attention mechanisms into CNNs by modeling global interactions of all pixels based on feature maps \cite{transunet}. The attention mechanism has been proposed to mimic the human visual system by concentrating portions of the most relevant information \cite{general_attention, medical_attention}. Attention mechanisms can be categorized into four groups: channel attention, spatial attention, hybrid channel-spatial attention, and branch attention. Squeeze-and-Excitation (SE) \cite{squeeze_excitation}, a channel attention method, exploits inter-channel dependencies using a squeeze operation followed by an excitation function. Convolutional Block Attention Module (CBAM) \cite{cbam} is a hybrid attention mechanism that applies attention to both spatial and channel dimensions. U-Net Attention \cite{unet_attention} employs Attention Gates (AGs) proposed by Oktay et al. to make the model attend to the pancreas in segmentation tasks.

\begin{figure*}[t]
\begin{center}
\includegraphics[width=1.0\linewidth]{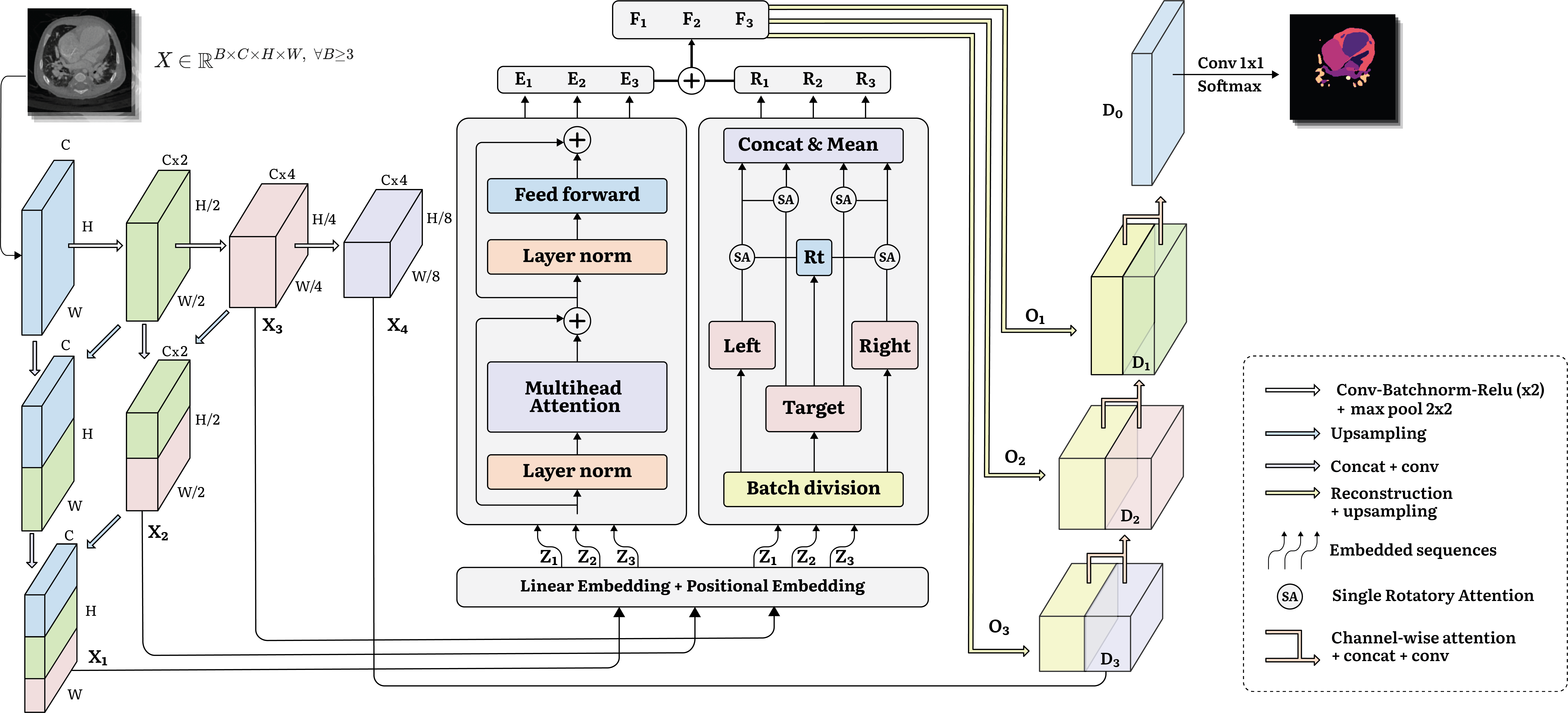}
\end{center}
\caption{RotCAtt-TransUNet++ architecture: Combining rotatory attention mechanism with channel-wise attention gates for enhanced feature fusion in decoder. Leveraging the Transformer-UNet Hybrid Model with enriched nested skip connections for multiscale feature extraction.}
\label{fig:network_architecture}
\end{figure*}

Channel-U-Net \cite{channel_unet} employs spatial channel-wise convolution to recalibrate spatial and channel-level features. SCAU-Net \cite{ScauNet} employs hybrid channel-spatial attention and integrates them as a plug-and-play module. Schlemper et al. \cite{additive_attention_gate} proposed additive attention gate modules which are integrated into skip connections. Despite attempts to integrate attention mechanisms into CNNs, these networks still have limitations. Inherent inductive biases limit CNN-based techniques from modeling long-range dependencies, while pooling and convolution layers might prevent low-level features from being propagated to the next convolutional layers. These architectures are intrinsically imperfect as they fail to exhibit long-range interactions and spatial dependencies, leading to a severe performance drop in the segmentation of medical images \cite{transnorm}. Additionally, these architectures generally yield weak performance, especially for target structures that exhibit large inter-patient variation in terms of texture, shape, and size \cite{transunet}.

\subsubsection{Transformer-based approaches}

In natural language processing (NLP), the ubiquitous architecture architecture of Transformer, designed for sequence-to-sequence prediction \cite{transunet}, has been seen as capable of learning long-term features \cite{transnorm}. Transformers were first proposed by \cite{attention_need} for machine translation, are not only significant at modelling global contexts but are also a promising tool for localizing local details \cite{transnorm}. The pioneering architecture, based purely on the self-attention mechanism, was the Vision Transformer (ViT) \cite{vit} which attained high performance compared to SOTA in image recognition tasks. Many cohort studies have investigated the combination of U-Net and Transformer to leverage both detailed high-resolution spatial information from CNN features and the global context encoded by Transformer. 

For example, TransUNet \cite{transunet} and UNetR \cite{unetr} employs Transformer as encoder to learn global information and CNNs as decoder to extract low-level spatial information. Theses networks utilize multiple self-attention heads to capture long-range dependencies. Above Transformer-CNN methods use the strategy of cutting input image into local patches (patchification), which raises two issues 'token-flatten issue' and 'scale-sensitivity issue' since Transformer flattens the local patches into $1D$ tokens, losing the interaction of tokenized information on local patches. Therefore, U-Netmer \cite{unetmer} was proposed to solve those 2 problems since it can segment input image with different patch sizes and by jointly training the U-Netmer, we can solve the scale sensitivity problem. Swin-Unet \cite{swinunet}, conversely, removes CNN and employs a complete Transformer architecture using shifted window mechanism to extract low-level details and a patch-expanding layer for upsampling. Attention Swin U-Net, with the enhanced skip connection by incorporation of attention mechanism into classical concatenation operation, was proposed for skin lesion segmentation. TransNorm employs the spatial normalization module from Transformer to enhance the decoder and skip connections. The Two-Level Attention Gate (TLAG) is also integrated. Azad at al. argued that expedient design of skip connections is crucial for accurate segmentation and achieved high accuracy with datasets International Skin Imaging Colloboration (ISIC) and Multiple Myeloma (MM) \cite{transnorm}. However, Transnorm still utilizes a skip connection between the bottleneck and the decoder paths, which can degrade the low-resolution information. In contrast, Attention Swin U-Net \cite{swinunet_attention} applies the attention mechanism in each encoder/decoder scale to model the multi-resolution feature representation. This network employs cross attention mechanism to enhance feature description on the skip connection path and imposes attention weights derived from encoder path to induce spatial attention mechanism, which achieves SOTA results on three public skin lesion segmentation datasets.

All the aforementioned Transformer-based approaches embed global self-attention with each patch representing a token. They share a commonality in that the attention mechanism is applied solely for interactions between patches or attention on the skip connection path. Additionally, these methods only process volumetric data slice by slice and can solely learn the interdependent interactions between patches in a single 2D image/slice. This limitation hinders TransUNet and its variants from extracting continuous information from adjacent slices, as evidenced by their fragmented structures after 3D reconstruction.

\subsubsection{3D and 2.5D networks}
While 3D networks like UNet 3D and VNet aim to retain inter-slice information along the z-axis, their practicality is hindered by high GPU memory requirements and computational costs during inference. On the other hand, 2.5D networks like AFTer-UNet aggregate information across slices, promising enhanced segmentation. However, AFTer-UNet lacks inter-slice attention and still demands substantial computational resources. 

In response to these challenges, we introduce RotCAtt-TransUNet++, a pioneering network merging Transformer-based and CNN-based architectures. With optimized nested downsampling and a unique rotatory attention mechanism, RotCAtt-TransUNet efficiently captures interslice connectivity while minimizing GPU memory usage and computational overhead. This innovative approach presents a novel pipeline for volumetric consideration in medical image segmentation.

\section{Methodology}
\label{sec:Methođology}
\begin{figure*}[t]
\begin{center}
\includegraphics[width=1.0\linewidth]{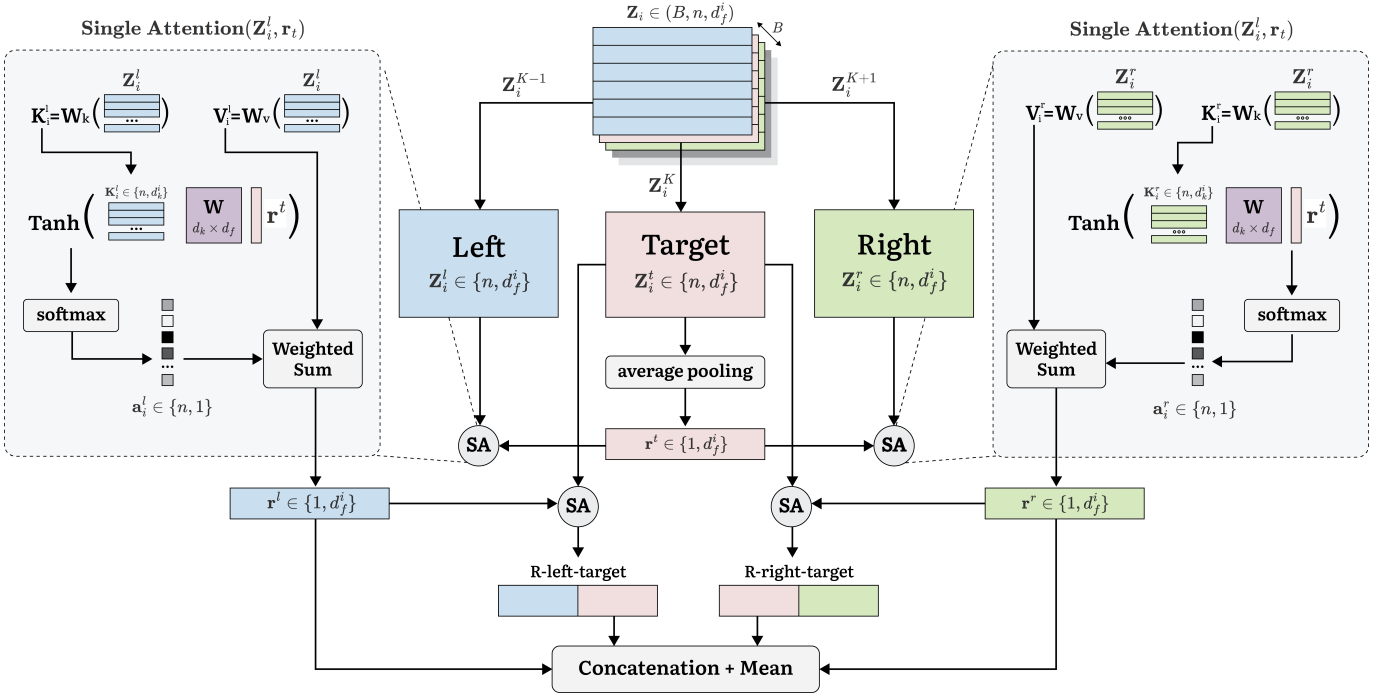}
\end{center}
\caption{The rotatory attention first uses the target phrase to compute new representations for the left (previous slice) and right (next slice) context using attention mechanism to capture the most important inter-connectivity information to current slice from two adjacent slices. Then, the second step use these left and right representations to calculate the new representations for the target phrase to integrate the most important information into the actual current slice itself.}
\label{fig:rotatory_attention_mechanism}
\end{figure*}

\subsection{Architecture Overview}
Through meticulous experimentation and ablation studies, we observed the efficacy of the UNet++ \cite{unet_plusplus} architecture coupled with dense downsampling and skip connections to preserve crucial information in achieving superior segmentation results. We are also inspired from pyramid pooling at different scales of Zhao at al \cite{pyramid}. Furthermore, \cite{transunet} also demonstrated that intensive incorporation of low-level features generally leads to a better segmentation accuracy. Thus, instead of the conventional CNN-based feature extraction approach, such as ResNet-50 in TransUNet, we embrace dense downsampling alongside nested skip connections, yielding four distinct feature maps $X_1, X_2, X_3, X_4$ at varying resolutions.

Unlike TransUNet and its variants which only embeds the last lowest-resolution feature maps, we employs linear embedding for multi-scale feature maps. Specifically, the first three feature maps $X_1, X_2, X_3$ undergo linear embedding with a different patch size $p$ to produce different embedded vector $z_i^j \in Z_i$, which simultaneously go through transformer blocks to capture the interactions between patches and rotatory attention mechanisms to aggregate the information from adjacent slices. Within these transformer blocks, comprising $N$ transformer layers, the embedded sequence patches traverse self-attention mechanisms and multilayer perceptrons, facilitating robust intra-slice information capture and yielding $E_1, E_2, E_3$.

The rotatory attention block, conceived to treat the batch size as a continuous slice, selectively processes three consecutive slices—designating the first as the left, the second as the target, and the third as the right—culminating in the production of $R_1, R_2, R_3$ encapsulating information from adjacent slices in the volumetric data. Integration of interslice and intraslice information yields $F_1, F_2, F_3$, which are then reconstructed to their original resolution via upsampling techniques, resulting in $O_1, O_2, O_3$.

Finally, $X_4$ undergoes concatenation with $O_3$, perpetuating this iterative process until the final segmentation map is obtained post $1 \times 1$ convolution.

\subsection{Multiscale Feature Extraction with Nested Shortcuts}
The input is structured as $(B, 1,H, W)$, representing the batch size, the number of channels (typically 1 in medical segmentation), height, and width, respectively. The batch size is also considered here since it also represents the number of adjacent slices whose information would be aggregated in rotatory attention block. This input undergoes convolutional operations to yield $X_1^1$, with a shape of $(B, C, H, W)$, where $C$ is set to 64 in our network. Subsequently, the resulting feature maps are downsampled to obtain $X_2^1$, with dimensions of $(B, C \times 2, \frac{H}{2}, \frac{W}{2})$. This $X_2^1$ is then upsampled to match the shape of $(B, C \times 2, H, W)$. Following this, $X_2^1$ and $X_1^1$ are concatenated along the $C$ axis, resulting in a shape of $(B, C \times 3, H, W)$, which then undergoes further convolution to produce $X_1^2$. This resultant tensor, $X_1^2$, shares the same shape as $X_1^1$ but encompasses aggregated information from $X_2^1$. This iterative process continues through subsequent lower resolution images. If we designate the desired number of different-resolution outputs as $D$, we then have $X_i^j$ $\space \forall i \in \{1, \dots, D-1\}$ and $\forall j \in \{1, \dots, D-i\}$, where $X_i^j$ has a shape of $(B, C \times 2^{i-1},\frac{H}{2^{i-1}}, \frac{W}{2^{i-1}})$. Notably, the $D$-th resolution map has a shape of $(B, C^{D-2}, \frac{H}{2^{D-1}}, \frac{W}{2^{D-1}})$, same depth as $D-1$-th resolution map and bypasses both the Transformer block and Rotatory Attention block but is instead utilized for the decoder. Specifically, when choosing $D = 4$, as in our case, the resulting feature maps are $X_1^3$, $X_2^2$, and $X_3^1$. For simplicity, these three $X$ tensors are denoted as $X_i$ for all $i \in \{1, 2, 3\}$. Subsequently, they are linearly embedded via convolution operations $E$ to produce patches represented as embedded vectors $z_j^{p_i} \in Z_i$ where $Z_i$ has shape $(B, n_i, d_f^i)$ and $1 \leq j \leq n_i$. The number of tokens or sequence length and the feature dimension of $Z_i$ are denoted as $n_i = {\frac{H_i \times W_i}{p_i^2}}$ and $d^f_i$ represent , respectively. Ensuring uniformity across $n_i$ for all $i$, we establish $D-1$ patch sizes $p_i = 2^{D-i+1}$, where $i$ ranges from 1 to $D-1$, implying that $p = \{2^4, 2^3, 2^2\}$ and the smallest patch size is $2^2 = 4$, given $D=4$. The multiscale feature extraction is illustrated in Figure 1 \ref{fig:network_architecture}.

Patch Embedding involves transforming vectorized patches $\hat{z}_j^{p_i} \in Z_i$ into a latent space of $d_i$ dimensions using a trainable linear projection. To preserve the spatial information of the patches, we incorporate position embedding specific to each patch, which are then combined with the patch embeddings.

\begin{gather*}
Z_i = E_i(X_i) + E_\text{pos}^i \\ 
Z_i = \hat{Z_i} + E^i_\text{pos} \\
[z_1^{p_i}, \dots, z_{n}^{p_i}] = [\hat{z}_1^{p_i}, \dots, \hat{z}_{n}^{p_i}] + [e^i_1, \dots, e^i_{n}]
\end{gather*}

where $E_i$ is the convolution operation to perform patch embbeding on $X_i$ and produce $\hat{Z}^i$, while $E_\text{pos}^i \in (B, n, d^i_f)$ denotes the position embedding, $Z_i$ is the linear embedding projection after adding vectors $\hat{z}_j^{p_i} \in (B, 1, d^i_f)$ with positional vectors $e_j^i \in (B, 1, d_f^i)$. The linear embedding and positional embedding is displayed in Figure 1 \ref{fig:network_architecture}.

\subsection{Transformer Block}
The Transformer encoder consists of $N$ layers of Multihead Self-Attention (MSA) and Multi-Layer Perceptron (MLP) blocks. Therefore the output of the $l$-th $\in N$ layer can be written as follows:

\begin{gather*}
\bar{Z}_i^{l'} = \text{MSA}(\text{LN}(Z_i^{l})) + Z_i^{l} \\
Z_i^{l+1} = \text{MLP}(\text{LN}(\bar{Z}_i^{l'})) + \bar{Z}_i^{l} \\
\cdots \\
\bar{Z}_i^{N-1} = \text{MSA}(\text{LN}(Z_i^{N-1})) + Z_i^{N-1} \\
Z_i^{N} = \text{MLP}(\text{LN}(\bar{Z}_i^{N-1})) + \bar{Z}_i^{N-1}
\end{gather*}

where $LN(\dot)$ denotes the layer normalization operator and $Z_i^l$ is the encoded image representation at scale $i$. The structure of Transformer layer is illustrated in Figure 1 \ref{fig:network_architecture}. In each layer $l$-th, the encoded image representation $Z_i$ undergo a self-attention mechanism, enabling encoded patches to learn how to attend to each other. Mathematically, the attention scores $A_i=\text{Attention}(Q_i,K_i,V_i)$ for $Z_i$ are computed as follows:

\begin{gather*}
A_i = \text{softmax}\left(\frac{Q_i K_i^T}{\sqrt{d_f^i}}\right) V_i
\end{gather*}

where $Q_i = W_q(Z_i), K_i = W_k(Z_i), V_i = W_v(Z_i)$ and $Q_i,K_i,V_i \in (B, n, d_f^i)$. Additionally, the Multilayer Perceptron contains a fully connected layer of size $d_i \times 4$ in the middle. The resulting $E_i$ maintains the same shape as $Z_i$, which learns the intraslice information or the relationship between patches in one 2D image slice.

\subsection{Rotatory Attention Block}
This technique is typically used in natural language processing, namely text sentiment analysis \cite{rot, general_attention} where there three inputs involved. The phrase for which the sentiment needs to be determined (target phrase), the text before target phrase (left context), text after target phrase (right context). This greedy method assumes that adjacent phrases would contribute the most to the current center/targer phrase. In our context, if we denote the current encoded input representation as $Z_i \in (B,n,d_f^i)$, we can separate this into multiple images $\{Z_i^1, \dots, Z_i^k, \dots, Z_i^B\}$. Therefore, three consecutive encoded slices/images can be selected as $\{ Z^{k-1}_i, Z^K_i, Z^{K+1}_i \}$ or $\{ Z^l, Z^t, Z^r \}$ to follow the left-target-right manner. For simple mathematical representation, we temporally disregard the scale $i$. These 3 encoded images are represented as:

\begin{gather*}
{ Z^l = [z_1^l, \dots, z_j^l, \dots, Z_{n}^l]} \in \mathbb{R}^{n \times d_f} \\
{ Z^t = [z_1^t, \dots, z_j^t, \dots, z_{n}^t]} \in \mathbb{R}^{n \times d_f} \\
{ Z^r = [z_1^r, \dots, z_j^r, \dots, z_{n}^r]} \in \mathbb{R}^{n \times d_f}
\end{gather*}

The idea is to extract a single vector $r \in d_f$ and add this vector to $Z^t$ to adjust the hidden states or transform the position of each embedded patch $z_j^t$ in the semantic dimensional space by referring to the information from adjacent slices. In detail, we need to represent $Z^t$ as a single vector $r^t$ and incorporate necessary information from left and right context by attention mechanism to avoid noise and redundant information. Firstly, a single target representation is created by using pooling layer that takes the average over rows of $Z^t$:

\begin{gather*}
{ r^t} = \text{pooling}(z^t_1, z^t_2, \dots, z^t_n) = \frac{1}{n} \Sigma_{j=1}^{n} {z^t_j}
\end{gather*}

Then similar to self-attention mechanism in Transformer layers, the key and value matrices are extracted from left context:

\begin{gather*}
K^l = W_k^l(Z^l) = [k_1^l, \dots, k^l_n] \in \mathbb{R}^{n \times d_f} \\
V^l = W_v^l(Z^l) = [v_1^l, \dots, v^l_n] \in \mathbb{R}^{n \times d_f}
\end{gather*}

The $r_t$ now is used as a query to create the context vector out of left context. The scores are calculated with activated general score function with $\text{tanh}$ activation function and the attention scores are calculated with softmax function: 
\begin{gather*}
S^l = [s^l_1, \dots, s^l_j, \dots, s_n^l] = \text{tanh}(K^l \cdot r^t + b^l) \\
a^l_j = \frac{\text{exp}({e^l_j})} {\Sigma_{j=1}^{n} \text{exp}(e^l_j)}
\end{gather*}

A weighted combination of patch embedding is considered as the component representation for left contexts:
\begin{gather*}
r^l = \Sigma_{i=1}^{n} a^l_i \cdot v^l_i
\end{gather*}

In Figure 2 \ref{fig:rotatory_attention_mechanism}, we denote the above process as Single Attention (SA), which is represented as:

\begin{gather*}
\text{SA}(Z, r) = 
\left\{
\begin{aligned}
  & K = W_k(Z), \quad V=W_v(Z) \\
  & a = \text{softmax}(\text{tanh}(K \cdot r + b)) \\
  & r = \sum_{n} a \cdot V
\end{aligned}
\right.
\end{gather*}

The vector $r^l$ is then used as query to create context out of target context to integrate information back into the center encoded slice/image to produce $r^{l/r} = SA(Z^t, r^l)$. An analogous procedure can be performed to obtain the right-aware target representation $r^r = SA(Z^r, r^t)$ and $r^{r/t} = SA(Z^t, r^r)$. Finally, to obtain the full representation vector $r$, we perform concatenation: $r^k = \text{concat}([{ r}^l, { r}^r, { r}^{l/t}, { r}^{r/t}])$ with $r^k \in \mathbb{R}^{1 \times d_f \times 4}$. This $r$ vector contains the aggregated information between 3 consecutive slices, thus we have $B-2$ vectors $r^k$ with $1 < k < B$ where $B$ is batch size since we perform the dense rotatory attention as illustrated in Figure 2 \ref{fig:rotatory_attention_mechanism}. The final vector $R$ is achieved as: $R = W_r(\text{mean}(r^k \space | \space 1 < k < B))$. But this is only one $i$-th level output, thus we have $R_i$ output. This interslice-informational vector is added to encoded intraslice-informational $E_i$ to retrieve more optimized vectorized patch embeddings $F_i$.

\subsection{Channel-wise Attention Gate for Feature Fusion}

\begin{figure}[t]
\begin{center}
\includegraphics[width=1.0\linewidth]{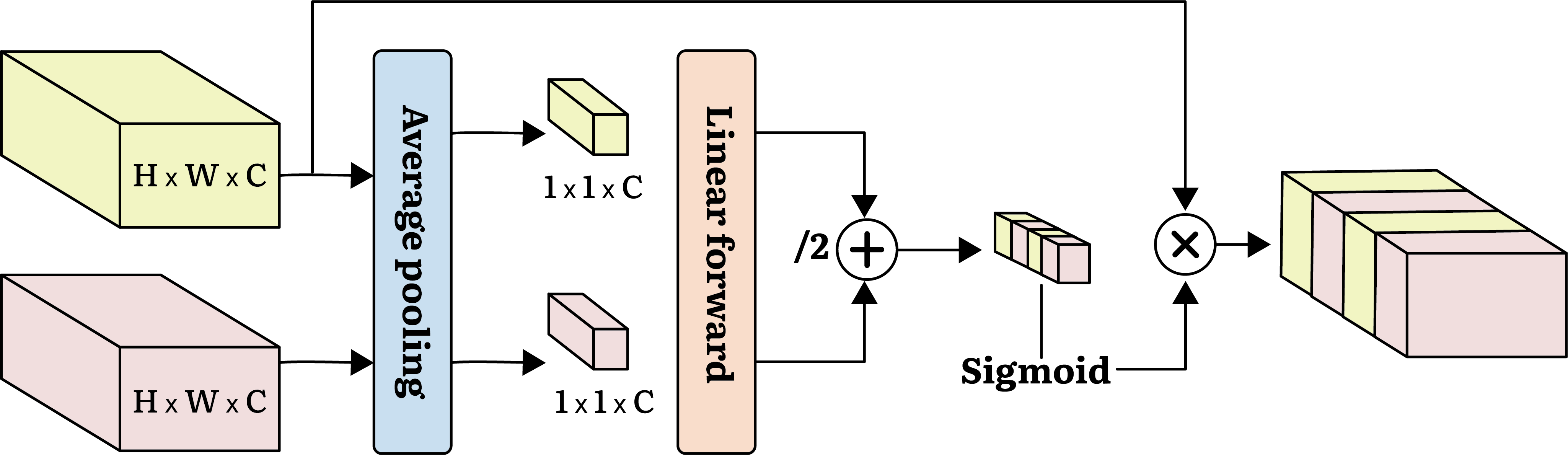}
\end{center}
\caption{The Channel-wise Attention Module integrates multi-scale context by incorporating cross attention from a channel-wise perspective. Its objective is to capture local cross-channel interactions, enabling an adaptive scheme for effectively merging multi-scale channel-wise features. This approach addresses potential scale semantic gaps through collaborative learning, rather than relying on independent connections, thereby resolving inconsistencies in semantic levels.}
\label{fig:predict_visualization}
\end{figure}

In order to better fuse features of inconsistent semantics between the Channel Transformer and U-Net decoder, we propose a channel-wise cross attention module, which can guide the channel and information filtration of the Transformer features and eliminate the ambiguity with the decoder features.

Mathematically, we take the $i$-th level output $F_i$ after Transformer and Rotatory blocks to reconstruct or decode the encoded image representations to get $O_i \in\mathbb{R}^{C \times H\times W}$. The reconstructed $O_i$ are taken with $i$-th level decoder feature map ${D_i}\in\mathbb{R}^{C \times H\times W}$ as the inputs of Channel-wise Cross Attention.

Spatial squeeze is performed by a global average pooling (GAP) layer, producing vector $\mathcal{G}({X}) \in\mathbb{R}^{C \times 1\times 1}$ with its $k^{th}$ channel $\mathcal{G}({X})=\frac 1 {H\times W} \sum_{i=1}^H\sum_{j=1}^W {X}^k(i,j)$. We use this operation to embed the global spatial information and then generate the attention mask:  

\begin{equation}
{M}_i={L}_1\cdot\mathcal{G}({O_i})+{L}_2\cdot\mathcal{G}({D_i})
\end{equation}

where ${L}_1\in\mathbb{R}^{C\times C}$ and ${L}_2\in\mathbb{R}^{C \times C}$ and being weights of two Linear layers and the ReLU operator $\delta (\cdot)$. This operation encodes the channel-wise dependencies. Followed ECA-Net \cite{eca} which empirically showed avoiding dimensionality reduction is important for learning channel attention, we use a single Linear layer and sigmoid function to build the channel attention map. The resultant vector is used to recalibrate or excite ${O}_i$ to ${\hat{O}}_i=\sigma({M}_i)\cdot {O_i}$, where the activation $\sigma({M}_i)$ indicates the importance of channels. Finally, the masked ${\hat{O}}_i$ is concatenated with the up-sampled features of the $i$-th level decoder.

\section{Experiments}
\label{sec:Experiments}
\begin{table*}[t]
\footnotesize
\begin{center}
\begin{threeparttable}
\caption{We evaluated the performance of architectures on five datasets, reporting three key metrics. The VHSCDD\textbf{*} dataset denotes images of size $512 \times 512$, while VHSCDD and other datasets (MMWHS, Synapse, ImageCHD) have images of size $256 \times 256$. TransUNet, was trained without utilizing a pre-trained Transformers model from ImageNet21K since we notice that incorporation of a pre-trained model did not yield significant performance improvements. Despite not being the most lightweight in terms of parameters, our architecture RotCAtt-TransUNet++ outperformed others across datasets and metrics, demonstrating its efficacy without relying on pre-trained Transformer models.}
{\jlrev
\label{tab:main_results}
\begin{tabular}{P{0.15\linewidth}P{0.05\linewidth}|
P{0.0265\linewidth}P{0.0265\linewidth}P{0.0265\linewidth}|
P{0.0265\linewidth}P{0.0265\linewidth}P{0.0265\linewidth}|
P{0.0265\linewidth}P{0.0265\linewidth}P{0.0265\linewidth}|
P{0.0265\linewidth}P{0.0265\linewidth}P{0.0265\linewidth}|
P{0.0265\linewidth}P{0.0265\linewidth}P{0.0265\linewidth}}
    \hline
    \multirow{2}*{\textbf{Architecture}} & \multirow{2}*{\textbf{Params}} &
    \multicolumn{3}{c|}{\textbf{MMWHS}}  & \multicolumn{3}{c|}{\textbf{Synapse}} & \multicolumn{3}{c|}{\textbf{ImageCHD}} & \multicolumn{3}{c|}{\textbf{VHSCDD}} & \multicolumn{3}{c}{\textbf{VHSCDD*}} \\
    
    & & \textbf{DSC} & \textbf{IOU} & \textbf{HD} & \textbf{DSC} & \textbf{IOU} & \textbf{HD} & \textbf{DSC} & \textbf{IOU} & \textbf{HD} & \textbf{DSC} & \textbf{IOU} & \textbf{HD} & \textbf{DSC} & \textbf{IOU} & \textbf{HD} \\
    \hline 
    UNet \cite{unet} & 124.2M & 0.78 & 0.61 & 28.3 & 0.61 & 0.43 & 30.5 & 0.72 & 0.52 & 26.1 & 0.50 & 0.29 & 39.4 & 0.449 & 0.26 & 89.5 \\
    Att-UNet \cite{unet_attention} & 32.54M & 0.84 & 0.78 & 15.6 & 0.51 & 0.33 & 44.9 & 0.86 & 0.75 & 20.2 & 0.40 & 0.23 & 42.9 & 0.51 & 0.34 & 92.1 \\
    UNet++ \cite{unet_plusplus} & 36.64M & 0.96 & 0.9 & 13.9 & 0.54 & 0.38 & 30.6 & 0.85 & 0.71 & 21.7 & 0.79 & 0.62 & 28.4 & 0.72 & 0.68 & 68.9 \\
    Att-UNet++ \cite{unet_plusplus_att} & 38.50M & 0.84 & 0.78 & 15.6 & 0.68 & 0.51 & 21.5 & 0.81 & 0.65 & 23.7 & 0.80 & 0.64 & 22.6 & 0.68 & 0.62 & 64.7 \\
    ResUNet \cite{resunet} & 52.17M & 0.76 & 0.64 & 17.6 & 0.47 & 0.31 & 40.6 & 0.68 & 0.56 & 34.2 & 0.56 & 0.35 & 41.9 & 0.61 & 0.56 & 40.9 \\
    \hline
    Swin-unet \cite{swinunet} & 165.4M  & 0.87 & 0.79 & 17.3 & 0.77 & 0.65 & \textbf{23.9} & 0.78 & 0.64 & 23.6 & 0.84 & 0.73 & 23.5 & 0.81 & 0.73 & 45.1 \\
    Att Swin-UNet \cite{swinunet_attention} & 165.4M &  0.84 & 0.73 & 20.4 & \textbf{0.79} & 0.67 & 24.5 & 0.89 & 0.78 & 18.7 & 0.82 & 0.71 & 25.6 & 0.79 & 0.65 & 43.1 \\
    TransUNet \cite{transunet} &  420.5M & 0.91 & 0.84 & 15.6 & 0.76 & \textbf{0.78} & 32.2 & 0.86 & 0.72 & 22.6 & 0.85 & 0.71 & 22.3 & 0.76 & 0.75 & 41.2 \\
    RotCAtt-TransUNet++ & 51.51M &  \textbf{0.97} & \textbf{0.92} & \textbf{15.9} & 0.68 & 0.61 & 25.6 & \textbf{0.96} & \textbf{0.89} & \textbf{15.67} & \textbf{0.93} & \textbf{0.91} & \textbf{20.3} & \textbf{0.95} & \textbf{0.92} & \textbf{32.4} \\
    \hline
\end{tabular}}
\end{threeparttable}
\end{center}
\end{table*}

\begin{figure*}[t]
\begin{center}
    \includegraphics[width=1.0\linewidth]{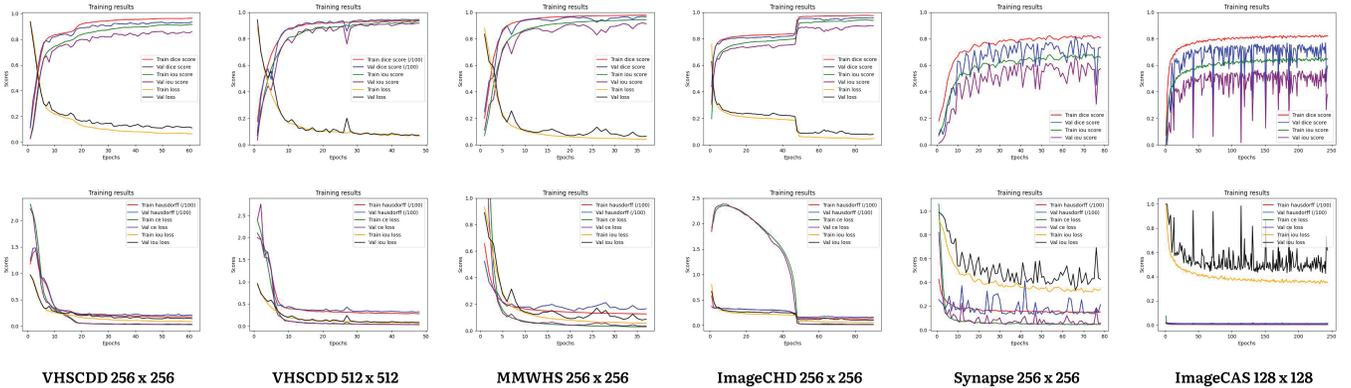}
    \end{center}
    \caption{The training graphs depict the performance of the RotCAtt-TransUNet++ model across five distinct datasets. Remarkably, our network excels when applied to cardiac data, benefiting from robust long-range interslice connectivity. However, we encountered challenges with the Synapse dataset, failing to meet anticipated performance levels. In case of ImageCAS, due to the dominance of background over coronary arteries in binary segmentation, our model exhibited limitations but still outperformed the baseline method (3D UNet) proposed by \cite{imagecas}}.
\label{fig:training_graphs}
\end{figure*}

\subsection{Datasets}
In our experimental phase, we delved into both binary segmentation and multi-class segmentation tasks across a diverse range of datasets divided into two types: one abdominal dataset and four cardiac datasets. Here's a detailed breakdown of the datasets used:

\subsubsection{Multi-Modality Whole Heart Segmentation Challenge 2017 (MMWHS-2017)}
The MMWHS-2017 dataset, sourced from the Multi-Modality Whole Heart Segmentation Challenge 2017 \cite{MMWHS}, comprises 20 MR and 20 CT volumes obtained from various clinical settings. For our experiments, we exclusively utilized the CT subset for training and validation. Expertly annotated by proficient individuals with backgrounds in biomedical engineering or medical physics, the dataset delineates seven fundamental cardiac regions: Left Ventricle (LV), Right Ventricle (RV), Left Atrium (LA), Right Atrium (RA), Myocardium of Left Ventricle (LV-Myo), Ascending Aorta Trunk (AA), and Pulmonary Artery Trunk (PA).

\subsubsection{Synapse multi-organ segmentation dataset}: We adopt a methodology akin to that employed by the authors of TransUNet \cite{transunet}, leveraging a dataset comprised of 30 abdominal CT scans sourced from the MICCAI 2015 Multi-Atlas Abdomen Labeling Challenge. These scans encompass a total of 3779 axial contrast-enhanced abdominal clinical CT images. Each CT volume spans a range of 85 to 198 slices, each measuring $512 \times 512$ pixels, with a voxel spatial resolution set at $([0.54 ~ 0.54] \times [0.98 ~ 0.98] \times [2.5 ~ 5.0])mm^3$. Following the methodology outlined in \cite{transunet}, our evaluation metrics include the average Dice Similarity Coefficient (DSC) and Hausdorff Distance (HD) computed across eight distinct abdominal organs: aorta, gallbladder, spleen, left kidney, right kidney, liver, pancreas, and stomach. To ensure the integrity of our performance comparison with TransUNet, we adhere to a consistent setup. This involves a randomized split of the dataset into 18 training cases, comprising 2212 axial slices, and 12 cases designated for validation. Notably, we utilize preprocessed data derived from TransUNet to maintain parity in our comparative analysis.

\begin{figure*}[t]
\begin{center}
    \includegraphics[width=1.0\linewidth]{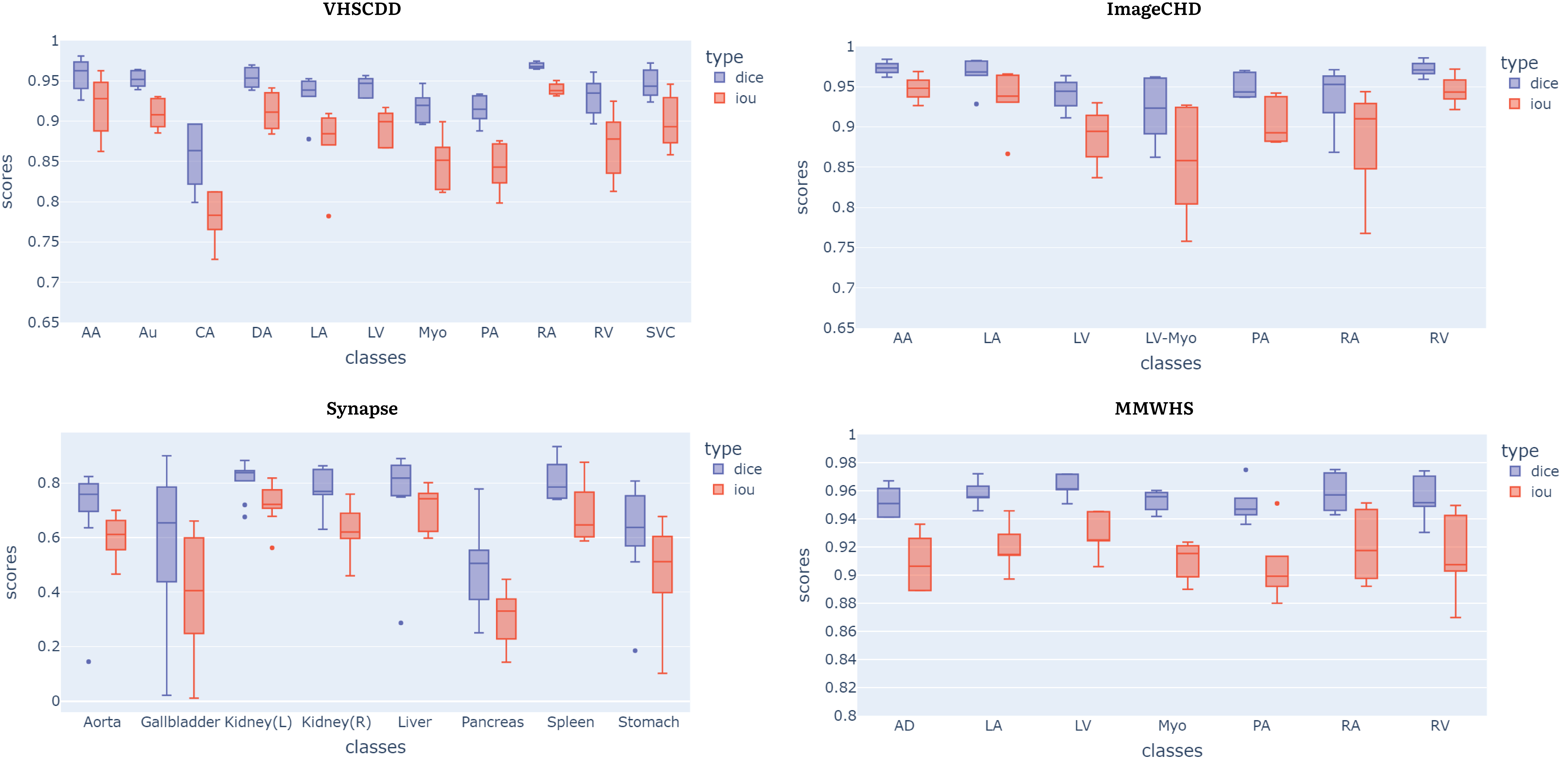}
    \end{center}
    \caption{Class-wise Dice Score and IoU scores of RotCAtt-TransUNet++ on VHSCDD, ImageCHD, Synapse, MMWHS datasets. Notably, CA (coronary arteries) exhibit the lowest scores (0.78-0.81), indicating a need for optimization. Moreover, myocardium, resembling background in CT scans, also shows low IoU scores across cardiac datasets. Compared to other architectures like TransUNet, our model demonstrates superior performance, addressing misprediction issues and avoiding "spraying phenomenon" in 3D reconstruction  (refer to \ref{fig:comparison}, \ref{fig:compare_vert}, and \ref{fig:rot}.)}
\label{fig:box_plot}
\end{figure*}

\subsubsection{ImageCHD - A 3D Computed Tomography}

The ImageCHD dataset \cite{imagechd} represents a significant resource for the classification of Congenital Heart Disease (CHD), comprising 110 3D Computed Tomography (CT) images. Notably, this dataset offers a nuanced labeling scheme, encompassing intricate details of cardiac small arteries and capillaries. With 8 segmented classes:Left Ventricle (LV), Right Ventricle (RV), Left Atrium (LA), Right Atrium (RA), Myocardium (Myo), Aorta duct (AD), Pulmonary Artery Trunk (PA), it provides a comprehensive view of the structural complexities inherent in CHD. Remarkably, ImageCHD features a diverse array of cases, encompassing 16 distinct congenital heart diseases alongside normal cases. This diversity extends to the shapes and sizes observed within specific cardiac regions, offering a rich dataset for analysis and classification tasks.Despite the dataset's complexity, the baseline methodology, employing UNet 3D and UNet 2D models with comparable configurations for training, yielded an average Dice Similarity Coefficient (DSC) of $75.6 \pm 10.2$. Notably, the segmentation performance varied across different cardiac structures, with great vessels exhibiting the lowest DSC at $66.5 \pm 15.1$, attributed to their intricate structures, while cardiac chambers achieved a higher DSC of $86.5$, owing to their clearer and more prominent shapes.

\subsubsection{ImageCAS - A Large-Scale Dataset and Benchmark for Coronary Artery Segmentation based on Computed Tomography Angiography Images}

\begin{table}[t]
\footnotesize
\begin{center}
\begin{threeparttable}
\caption{We investigated the impact of varying input image sizes while maintaining fixed patch sizes, denoted as $p_i \in \{16, 8, 4\}$. Consequently, as the input image size increases by a factor of $2$, the number of tokens increases by $4$ times. However, excessively small image sizes may lead to fragmented segmentation maps and 3D reconstructed structures. Additionally, we analyze the influence of the number of Transformer layers (TLs). Surprisingly, we observe that the number of layers does not significantly affect performance. However, intriguingly, we find that setting $\text{TLs}=4$ yields the best results on the VHSCDD dataset, even with the same 60 epochs of training.}

{\jlrev
\label{tab:ablation_rot}
\begin{tabular}
{P{0.04\linewidth}P{0.03\linewidth}P{0.08\linewidth}P{0.13\linewidth}P{0.13\linewidth}P{0.13\linewidth}P{0.13\linewidth}}
    \hline
    {\textbf{Size}} & {\textbf{TLs}} & {\textbf{Params}} & {\textbf{DSC}} & {\textbf{IOU}} & {\textbf{HD}} & {\textbf{CE}} \\
    \hline
    $128$ & 4 & 51.51M & 0.916{\tiny $\pm$0.061} & 0.842{\tiny $\pm$0.054} & 14.265{\tiny $\pm$1.65} & 0.038{\tiny $\pm$0.16}\\
    $256$ & 4 & 51.51M & 0.927{\tiny $\pm$0.042} & 0.894{\tiny $\pm$0.037} & 20.263{\tiny $\pm$1.21} & \textbf{0.032}{\tiny $\pm$0.12}\\
    $256$ & 9 & 70.55M & \textbf{0.934}{\tiny $\pm$0.041} & \textbf{0.911}{\tiny $\pm$0.043} & \textbf{18.878}{\tiny $\pm$1.38} & 0.035{\tiny $\pm$0.14}\\
    \hline
    $512$ & 3 & 47.71M & 0.904{\tiny $\pm$0.078} & 0.916{\tiny $\pm$0.081} & \textbf{31.983}{\tiny $\pm$1.89} & 0.042{\tiny $\pm$0.24}\\
    $512$ & 4 & 51.51M & \textbf{0.945}{\tiny $\pm$0.052} & \textbf{0.918}{\tiny $\pm$0.067} & 32.380{\tiny $\pm$1.59} & \textbf{0.035}{\tiny $\pm$0.18}\\
    $512$ & 9 & 70.55M & 0.919{\tiny $\pm$0.069} & 0.905{\tiny $\pm$0.076} & 33.019{\tiny $\pm$1.78} & 0.043{\tiny $\pm$0.24}\\
\hline
\end{tabular}}
\end{threeparttable}
\end{center}
\end{table}

This is a comprehensive dataset \cite{imagecas} comprising 3D CTA images obtained using a Siemens 128-slice dual-source scanner, encompassing data from 1000 patients. Among these patients, those previously diagnosed with coronary artery disease and who underwent early revascularization are included in the dataset. Each image measures $512 \times 512$ pixels with $206$ to $275$ axial slices per volume. The images boast a planar resolution ranging from $0.29$ to $0.43 mm^2$, with a slice spacing of $0.25$ to $0.45 mm$. Originating from authentic clinical scenarios at the Guangdong Provincial People's Hospital, this dataset serves for binary segmentation purposes. However, challenges arise as the background area within a slice often overwhelms the coronary arteries, leading to fragmented segmentation and reconstruction. Given the elongated nature of coronary structures along the z-axis, the author \cite{imagecas} implemented a 3D UNet approach. Yet, direct segmentation of the entire 3D image at its original resolution proves infeasible due to substantial memory requirements. Consequently, the author adopted supplementary techniques, such as coarse segmentation on lower-resolution images and skeleton extraction. Despite these efforts, the achieved Dice Similarity Coefficient (DSC) remains relatively modest; specifically, the DSC for a $128 \times 128$ resolution hovers around $0.68$.

\subsubsection{VHSCDD: Vietnamese Heart Segmentation and Cardiac Disease Detection}

The data acquisition process involved capturing raw CT/CTA slice images using the Toshiba Aquilion ONE CT scanner, sourced from patient scans. Annotation was conducted across 12 classes (one backround): left ventricle, right ventricle, left atrium, right atrium, descending aorta, aortic arch, vena cava, pulmonary trunk, myocardium, coronary arteries, and auricle. Drawing inspiration from the meticulously annotated ImageCHD dataset, we leveraged models trained on ImageCHD to predict labels for new raw data sourced from reputable hospitals across Vietnam. Subsequently, we refined the segmentation results, placing particular emphasis on enhancing annotations for coronary arteries, the auricle, and the vena cava. 

\begin{figure}[t]
\begin{center}
\includegraphics[width=1.0\linewidth]{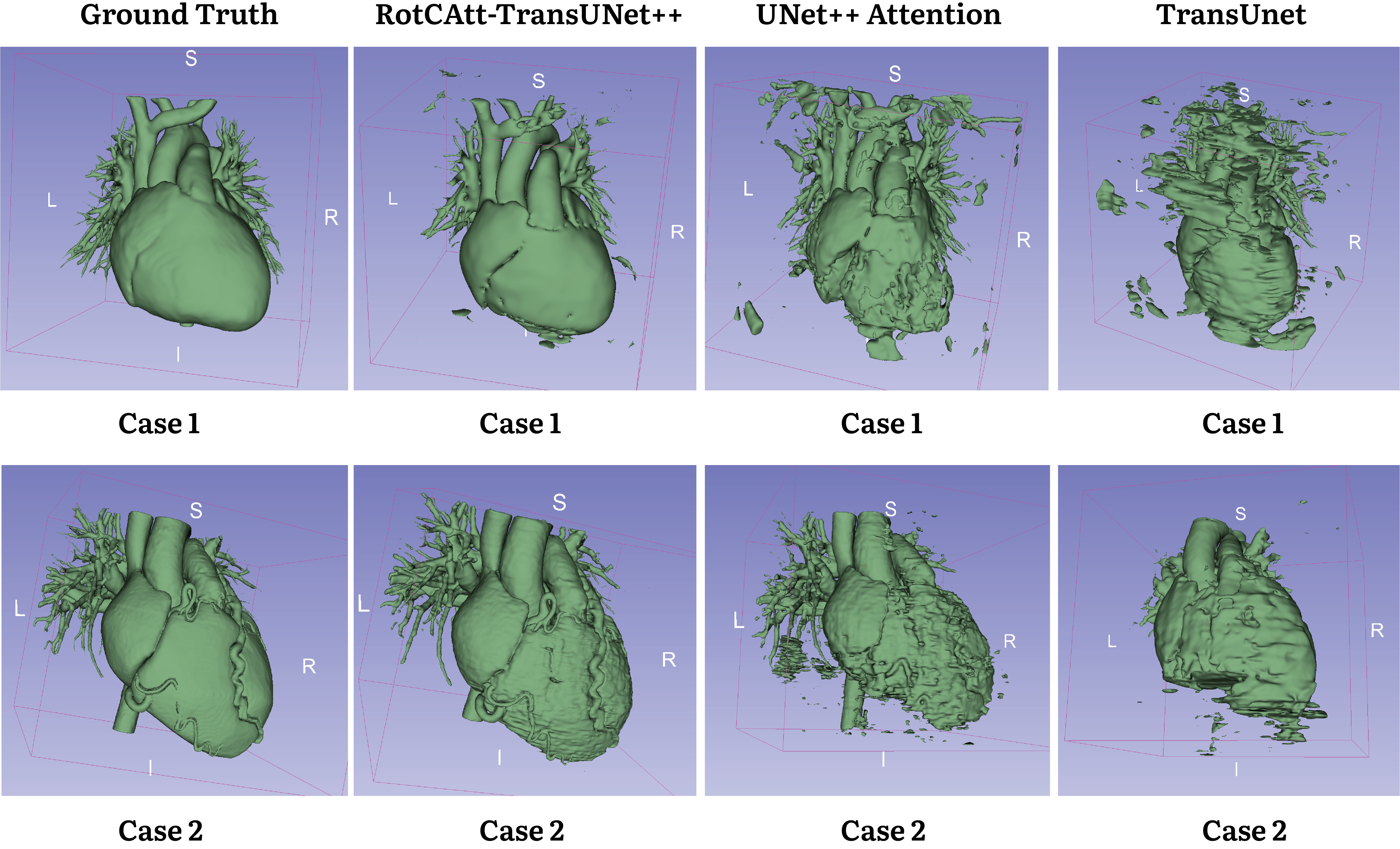}
\end{center}
\caption{Comparing 3D reconstructions from our model with TransUNet and UNet++ Attention: TransUNet exhibits a 'spraying' phenomenon, while UNet++ Attention tends to overlook crucial details.}
\label{fig:comparison}
\end{figure}

The VHSCDD dataset stands out for its exceptional level of detail, particularly in delineating intricate vascular structures such as small arterioles and arteries. This granular level of annotation presents a novel challenge for state-of-the-art (SOTA) algorithms, as existing approaches often struggle to achieve satisfactory Dice Similarity Coefficients (DSC) for classes like coronary arteries and the auricle. Additionally, distinguishing between the background and myocardium poses a notable challenge due to their visual similarity. Comprising 56 volumetric 3D cases, the VHSCDD dataset features images with dimensions $512 \times 512 \times 35-450$. We experimented at different resolutions, including $128 \times 128$, $256 \times 256$, and $512 \times 512$ with fixed patch sizes of slices only in axial view.

\subsection{Implementation details}

\begin{figure}
\begin{center}
\includegraphics[width=1.0\linewidth]{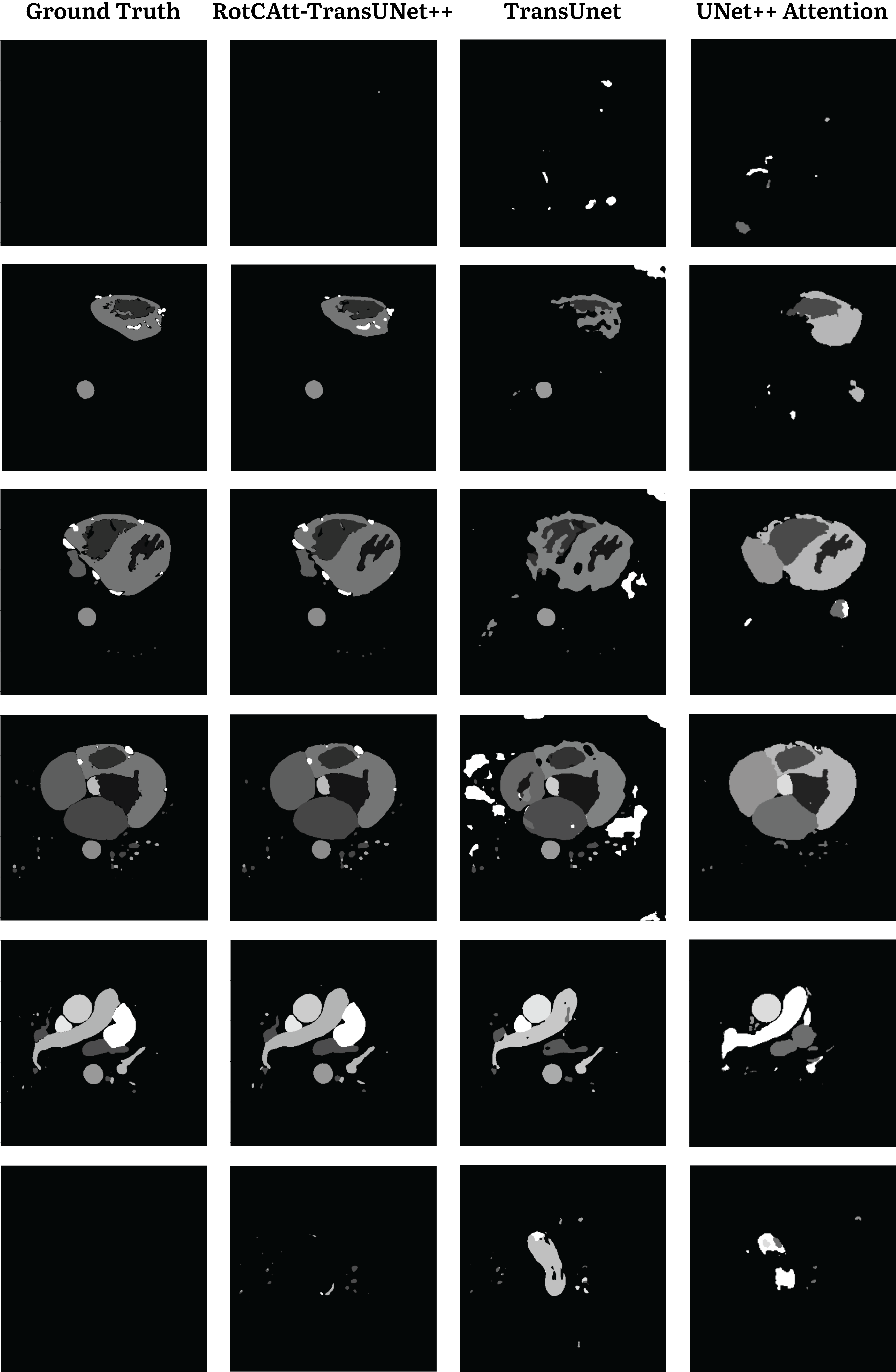}
\end{center}
\caption{Comparing 2D segmentation among our model, TransUNet, and UNet++ Attention for case $2$, consisting of 400 slices, reconstructed in 3D (see Figure \ref{fig:comparison}). Our model predicts based on batch size steps, whereas the others predict slice by slice. Upon scrutiny, while our model's segmentation isn't identical to the label, it closely approximates it, which is acceptable. In contrast, the results from the other models fail to meet the standard.}
\label{fig:compare_vert}
\end{figure}

We used NVIDIA RTX 4090 1X GPU with 24GB memory, 81.4 TFLOPS for the training process. For our experiments, we utilized the NVIDIA RTX 4090 1X GPU, with 24GB of memory, 81.4 TFLOPS for our training tasks. We implemented our network RotCAtt-TransUNet++ with 8 different networks: TransUNet, Swin-unet, Attention Swin-UNet, UNet, UNet Attention, UNet++, UNet++ Attention, ResUNet. Across 5 diverse datasets, we evaluated the performance of 9 different networks using essential metrics such as Dice Coefficient Score (DSC), Intersection over Union (IoU) scores, and Hausdorff Distance. For a detailed class-wise analysis, we provided supplementary class-wise DSC and class-wise IoU scores. Nine networks were implemented using PyTorch, employing a fixed configuration for patch size $p_i$ and embedding dimension $d^i_f$, where $i$ signifies distinct feature map scales. Specifically, we utilized $p = [16, 8, 4]$ and $d_f = [64, 128, 256]$. Consequently, for input image sizes of $128, 256, 512$, we had token counts of $64, 256, 1024$ respectively. Additionally, we saved the matrices of self-attention weights, context weights, and rotatory attention vectors denoted as $A, C, R$ respectively, for visualization and ablation study purposes. We consciously avoided employing any data augmentation techniques to maintain the synthetic nature of our data and to prevent the introduction of extraneous artifacts that could potentially bias performance comparisons between models.For optimization, we chose the Stochastic Gradient Descent (SGD) optimizer with an initial learning rate set at $0.01$ and a weight decay of $0.0001$. However, our code implementation also provides an option for the Adam optimizer.

We employed a 3 loss functions: Cross Entropy Loss, Dice Loss, and IoU Loss, leveraging the combined loss of Dice Coefficient (DSC) and Intersection over Union (IoU) for efficient backpropagation. The mathematical formulations are as follows:
$$
CE = -\frac{1}{N} \sum_{i=1}^{N} \sum_{j=1}^{C} G_{ij} \log(P_{ij} + (1 - G_{ij}) \log(1 - P_{ij} )
$$

Cross Entropy Loss quantifies the disparity between the predicted probability distribution ($P_{ij}$) and the ground truth labels ($G_{ij}$). It calculates the average negative logarithm of the predicted probabilities assigned to the correct classes. This loss function is commonly employed in classification tasks to guide the model towards minimizing classification errors.

\begin{figure}
\begin{center}
\includegraphics[width=1.0\linewidth]{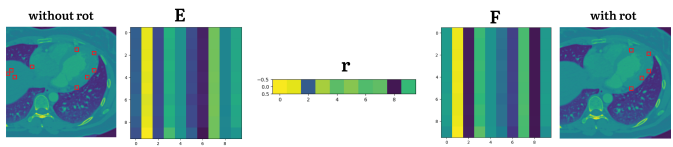}
\end{center}
\caption{Utilizing a rotatory attention mechanism, we transform the encoded image representation $E$ into $F$, aimed at averting distractions from non-cardiac details in chest CT scans to enhance myocardium segmentation, thus mitigating the 'spraying' phenomenon and facilitating refined segmentation and 3D reconstruction. In row $E$, each vectorized embedded patch represents semantically dimensional features, with each column denoting specific features. The brighter color (e.g. yellow) indicates the focused feature. Notably, $F$ retains focus on the most crucial feature while adjusting other feature values.}
\label{fig:rot}
\end{figure}

$$
\text{Dice Loss} = 1 - \frac{2 \sum_{ij} P^c_{ij} \times G^c_{ij}}{\sum_{ij} P^c_{ij} + \sum_{ij} G^c_{ij} + \epsilon} \quad \forall c \neq 0
$$
Dice Loss measures the dissimilarity between the predicted segmentation ($P$) and the ground truth ($G$) by computing the Dice coefficient. It assesses the overlap between the two sets, emphasizing regions of agreement while penalizing inconsistencies. This loss is particularly effective in scenarios where class imbalances exist, as it provides a robust measure of segmentation accuracy.
$$
\text{IoU Loss} = 1 - \frac{\sum_{ij} P^c_{ij} \times G^c_{ij}}{\sum_{ij}(P^c_{ij} + G^c_{ij} - P^c_{ij} \times G^c_{ij})} \quad \forall c \neq 0
$$

IoU Loss, or Intersection over Union Loss, evaluates the spatial overlap between the predicted and ground truth segmentation masks. It quantifies the ratio of the intersection area to the union area of the two sets, providing a comprehensive measure of segmentation accuracy. By penalizing deviations from ideal overlap, IoU Loss guides the model towards producing segmentation maps that closely align with ground truth annotations.

Here, $P$ and $G$ represent the predicted segmentation map and ground truth respectively, while $c$ denotes the class. The exclusion of $c \neq 0$ ensures the avoidance of unreal DSC and IoU scores stemming from dominant background pixels. Our composite loss function is defined as:

\section{Results and Conclusion}
\label{sec:ResultsAndConclusion}

$$
L = \alpha \times \text{IoU Loss} + (1 - \alpha) \times \text{Dice Loss}
$$

In our implementation, we set $\alpha$ to $0.6$ to balance the contributions of both losses effectively. Additionally, we compute the Hausdorff distance:

$$
HD(P, G) = \max\left(\max_{p \in P} \min_{g \in G} \| p - g \|_2, \max_{p \in P} \min_{g \in G} \| p - g \|_2\right)
$$

Here, $\| p - g \|_2$ denotes the Euclidean distance between points $p$ and $g$. This metric provides valuable insights into the dissimilarity between two sets of points, aiding in evaluating the effectiveness of our segmentation approach.

The validation results of 9 models across various datasets are presented in Table \ref{tab:main_results}. Additionally, the training graphs and the class-wise DSC and IoU scores of our model across datasets are displayed in Figure \ref{fig:training_graphs} and Figure \ref{fig:box_plot}, respectively. We conducted an ablation study on different input image sizes and varying numbers of Transformer layers, as shown in Table \ref{tab:ablation_rot}. The 2D segmentation results and 3D reconstruction, presented in Figure \ref{fig:comparison} and Figure \ref{fig:compare_vert}, respectively, showcase our model's performance compared to Transformer-based method (TransUNet) and CNN-based methods (UNet++ Attention). Furthermore, we interpret the results by visualizing the interaction between patches and the encoded image representation in Figure \ref{fig:rot}.

\begin{figure}
\begin{center}
\includegraphics[width=1.0\linewidth]{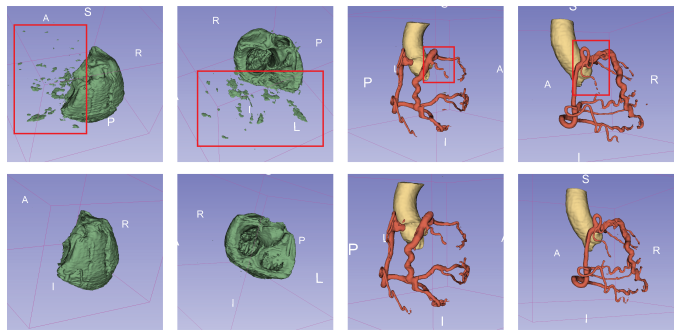}
\end{center}
\caption{Delving deeper into the impact of the Rotatory Attention Mechanism on altering semantically dimensional features within vectorized patches to attain superior segmentation refinement and optimize 3D reconstruction.}
\label{fig:3D_rot}
\end{figure}

\begin{figure}
\begin{center}
\includegraphics[width=1.0\linewidth]{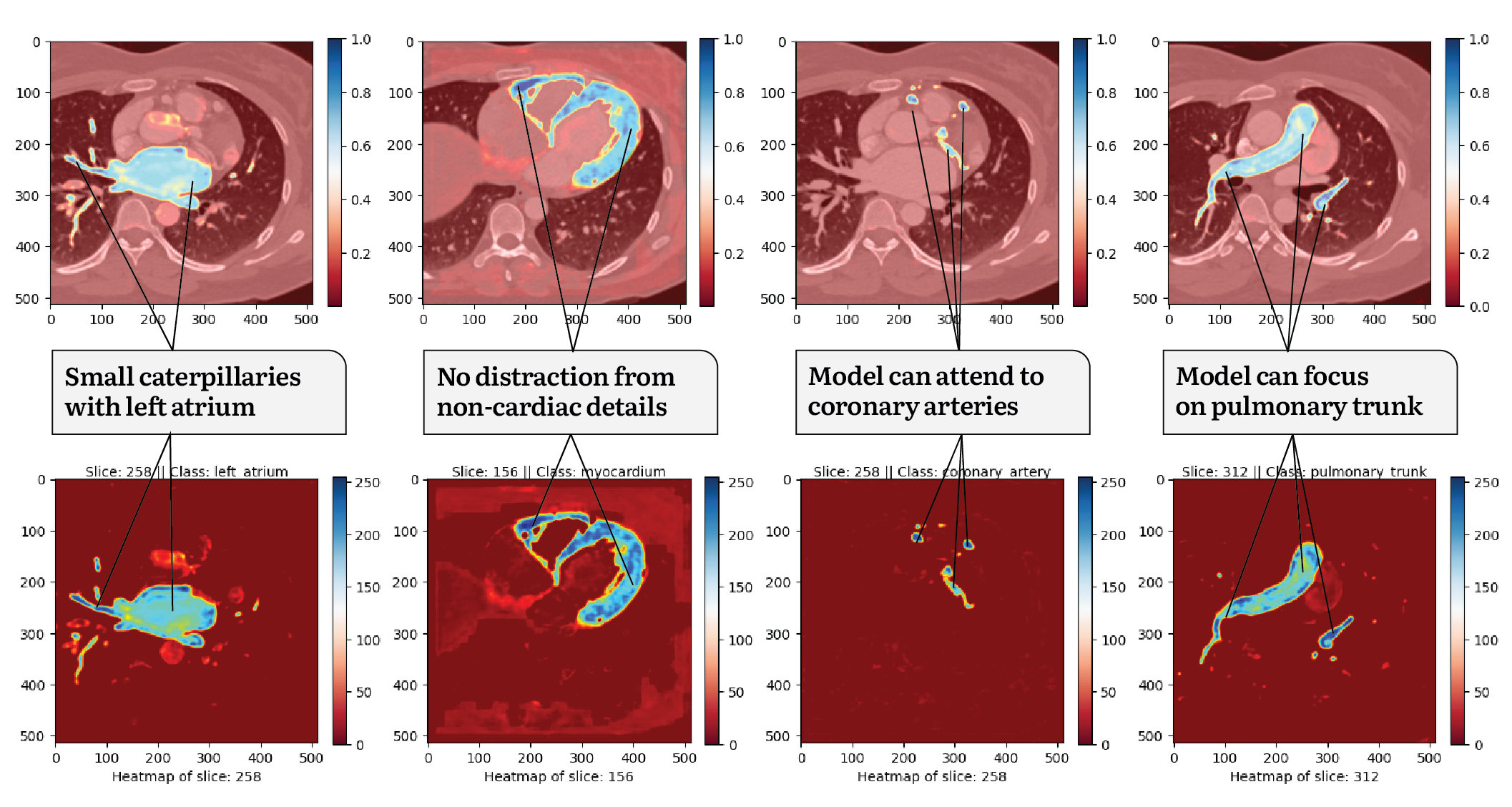}
\end{center}
\caption{Intepretable model: To analyze the specific regions that our model focuses on during segmentation, we adapted the GradCam algorithm for medical segmentation tasks. The blue-colored regions represent where model attends highly, while the red-colored ones are the regions of low attention. The heatmap visualization reveals that our model accurately targets the most relevant areas across all 12 classes. Specially, the minuscule details such as coronary artery are not ignored but accurately segmented, while other non-cardiac structures are not mistaken with myocardium.}
\label{fig:gradcam}
\end{figure}

In conclusion, Transformer-based methods are recognized for their robust innate self-attention mechanism, whereas CNN-based methods demonstrate proficiency in localization tasks. The most prominent and recent model, TransUNet, still exhibits limitations in capturing inter-slice information, thereby impeding intra-slice information capture as well. our study introduces RotCAtt-TransUNet++. a novel architecture that integrates nested skip connections and dense downsampling for multi-scale feature extraction in the encoder, followed by obtaining multi-scale feature maps through transformer layers and rotatory attention blocks. This process yields a better encoded image representation, utilized in the decoder path for accurate segmentation map reconstruction. our model achieves superior segmentation accuracy, particularly in datasets featuring complex cardiac structures. Experimental results across multiple datasets demonstrate near-perfect annotation of critical structures like coronary arteries and myocardium, underscoring the model's efficacy in real-world scenarios. The ablation study further validates the effectiveness of the rotatory attention to improve segmentation accuracy and efficiency. Further research contributes to automating medical image segmentation, reducing manual annotation burdens, and facilitating timely diagnosis of cardiovascular diseases. 

\bibliographystyle{IEEEtran}
\bibliography{references}
\end{document}